\documentclass[11pt]{article}

\usepackage[final]{acl}

\usepackage{times}
\usepackage{latexsym}

\usepackage[T1]{fontenc}

\usepackage[utf8]{inputenc}

\usepackage{microtype}

\usepackage{inconsolata}

\usepackage{graphicx}

\title{\method{}: Multi-Agent Web Navigation via Global-View Optimization}

\author{Weixi Tong \\
  Purdue University \\
  \texttt{tong172@purdue.edu} \\\And
  Yifeng Di \\
  Purdue University \\
  \texttt{di5@purdue.edu} \\\And
  Tianyi Zhang \\
  Purdue University \\
  \texttt{tianyi@purdue.edu}
  }

\usepackage{hyperref}
\usepackage{algorithm}
\usepackage{algpseudocode}
\usepackage{amsmath}
\usepackage{longtable}
\usepackage{array} 
\usepackage{booktabs}
\usepackage{colortbl}
\usepackage{booktabs}
\usepackage[table]{xcolor}
\usepackage{multirow}
\usepackage{tabularx}
\usepackage{subcaption}
\usepackage{CJKutf8}

\definecolor{sagegreen}{RGB}{225, 235, 225}
\definecolor{dustyblue}{RGB}{220, 235, 245}

\newcommand{\headerblue}{\rowcolor{dustyblue}}
\newcommand{\headergreen}{\rowcolor{sagegreen}}
\newcommand{\headercolorlong}{\rowcolor{gray!17}}
\newcommand{\method}{\textsc{Mango}}

\begin{document}
\maketitle
\begin{abstract}
Existing web agents typically initiate exploration from the root URL, which is inefficient for complex websites with deep hierarchical structures. Without a global view of the website's structure, agents frequently fall into navigation traps, explore irrelevant branches, or fail to reach target information within a limited budget. We propose \method{}, a multi-agent web navigation method that leverages the website structure to dynamically determine optimal starting points. 
We formulate URL selection as a multi-armed bandit problem and employ Thompson Sampling to adaptively allocate the navigation budget across candidate URLs. Furthermore, we introduce an episodic memory component to store navigation history, enabling the agent to learn from previous attempts. Experiments on WebVoyager demonstrate that \method{} achieves a success rate of 63.6\% when using GPT-5-mini, outperforming the best baseline by 7.3\%. Furthermore, on WebWalkerQA, \method{} attains a 52.5\% success rate, surpassing the best baseline by 26.8\%. We also demonstrate the generalizability of \method{} using both open-source and closed-source models as backbones. Our data and code are open-source and available at \url{https://github.com/VichyTong/Mango}.

\end{abstract}

\section{Introduction}

There has been growing interest in building web agents using large language models (LLMs)~\cite{zhou2024webarena, wang2023voyager, wu-etal-2025-webwalker, akkil2024agente, yang2025agentoccam}. %
These agents typically start from the root URL (i.e., the homepage) of a website, navigate and parse web content, and interact with the browser through actions such as clicking buttons and typing into text fields. 

Given the complexity of real-world websites, the root URL of a website may not be an optimal starting point. These agents have to traverse the website's structure from the top down, often exploring irrelevant sub-trees, falling into navigation traps, or wasting computation on generic intermediate pages. While prior work aims to enhance web navigation efficiency via better decision-making~\cite{abuelsaad2024agent, tan2025cradle} or alignment~\cite{kim-etal-2024-auto, yang2025agentoccam}, such methods are still limited by partial observations at each step without a global view of the website's structure.  

To address this limitation, we propose \method{}, a \textbf{M}ulti-\textbf{A}gent \textbf{N}avigation method with \textbf{G}lobal-view \textbf{O}ptimization for web automation. 
Given a target website, \method{} first constructs the global structure of the website through a lightweight crawl and a site-specific keyword search. Building upon this structure, \method{} further identifies a candidate set of URLs that are most relevant to the user query. Given a limited budget, \method{} dynamically selects the next URL to visit via Thompson Sampling. After each navigation attempt, \method{} reflects on the navigation trajectory and updates the probabilistic distribution of Thompson Sampling. It also stores both the reflection and the trajectory in an episodic memory to prevent redundant visits.

\method{} is evaluated on two web navigation benchmarks: WebVoyager~\cite{he-etal-2024-webvoyager} and WebWalkerQA~\cite{wu-etal-2025-webwalker}. Following prior works on web navigation~\cite{wu-etal-2025-webwalker, yang2025agentoccam}, we use success rate (SR) and action count (AC) to measure the performance and efficiency of \method{}. We experiment with five LLMs as backbones for \method{} and compare it against two state-of-the-art (SOTA) methods, AgentOccam~\cite{yang2025agentoccam} and WebWalker~\cite{wu-etal-2025-webwalker}, on both benchmarks. The results show that \method{} consistently achieves superior SR compared to SOTA baselines across all five backbone models, yielding absolute improvements of 3.1\% to 7.3\% on WebVoyager and 4.6\% to 26.8\% on WebWalkerQA. 
Regarding efficiency, \method{} maintains competitive or lower AC on the Qwen3 models. While it incurs higher AC when using GPT-5-mini, this investment translates into performance gains, yielding a 7.3\% absolute improvement on WebVoyager and a 26.8\% absolute improvement on WebWalkerQA compared to the best baseline. Finally, ablation studies  show that \method{} significantly outperforms variants based on random URL selection, search-based URL selection, and MCTS-guided navigation, which confirms the effectiveness of global view analysis and Thompson Sampling used in \method{}. 

\section{Related Work}

\subsection{Web Navigation Agents}
\label{sec: llm-based-web-navigation-agents}

Recent advances in LLMs have significantly accelerated research on web navigation agents~\cite{ning2025survey}.
One line of research focuses on improving agents' perception of the browser environment~\cite{deng2023mindweb, zheng2024synapse}, e.g., by summarizing long HTML documents into task-relevant snippets to extract salient environmental information~\cite{gur2024real}.
Another line of research focuses on planning and decision-making, where complex tasks are decomposed into a sequence of sub-tasks that are executed step by step~\cite{li-etal-2023-zero, abuelsaad2024agent, tan2025cradle}.
A third line of research aims to refine action execution through better grounding in the browser environment, e.g., aligning natural language actions with correct UI elements~\cite{lin2025showui, kim-etal-2024-auto}. 

However, these methods share a common limitation: they typically initiate navigation from the root URL (i.e., the homepage), leading to inefficient exploration on large and complex websites. In contrast, our method explicitly constructs a global structural representation of the website to identify intent-related entry points for navigation.

\subsection{Agentic Search Strategies}

To enhance the multi-step reasoning capability of agents, different search strategies have been adopted to navigate the solution space of a task in different domains~\cite{gan-etal-2025-master, antoniades2025swesearch, yu2025exact}.
For instance, Language Agent Tree Search~\cite{zhou2024language} integrates Monte Carlo Tree Search (MCTS) with LLM-powered value functions and self-reflection to optimize decision-making in the generation process. In web navigation,  WebPilot~\cite{zhang2024webpilot} adopts MCTS with a dual optimization strategy.
\citet{koh2025tree} adopts a best-first search algorithm to explore diverse interaction trajectories for improved task completion.

While effective, these methods still must discover the website structure incrementally from the homepage. Our method differs by pruning the search space before navigation begins. Instead of expanding a massive search tree from the root, we use the global structure of the website to identify candidate URLs and then use Thompson Sampling to allocate the computational budget efficiently. As shown in Section~\ref{sec:ablation study}, \method{} consistently outperforms the MCTS variant, achieving absolute improvements ranging from 5.2\% to 17.1\%.

\subsection{Agentic Memory}
\label{sec: agetic_memory}
Agentic memory is an effective way to transform stateless LLMs into adaptive agents by enabling the autonomous retention, organization, and retrieval of experiential knowledge for long-horizon planning and continuous adaptation~\cite{xu2025amem}. Some web agents adopt short-term memory, which stores previous actions in a web navigation trajectory~\cite{guan2024intelligent, Lai2024autowebglm}.
\citet{zheng2024synapse} propose a method to store and retrieve web navigation trajectories of successfully completed tasks to guide the current navigation.
\citet{agashe2025agent} introduced Narrative Memory, which stores high-level summaries of past experiences and real-time information retrieved from the web.

Building on these foundations, we integrate an episodic memory module to prevent the agent from repeating unsuccessful actions. \method{} records the actions performed in each navigation trajectory and its reflection on the trajectory. 
During subsequent navigation to the same URL, the stored information is provided to the navigation agent to prevent it from repeating incorrect actions.

\section{Method}
\begin{figure*}[tbp]
  \centering
  \includegraphics[width=0.98\linewidth]{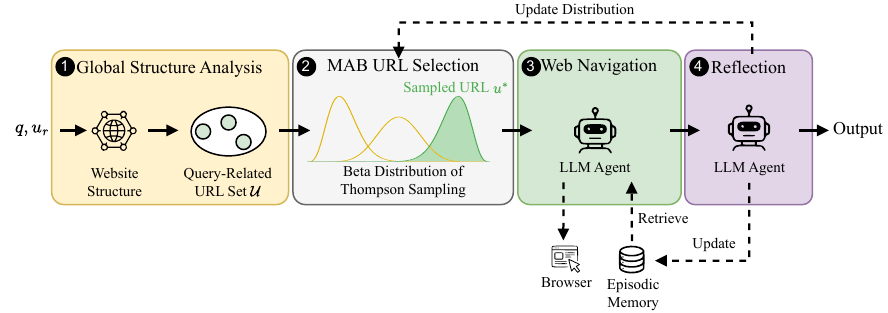}
  \caption{Overview of the web navigation agent system.}
  \label{fig:system}
\end{figure*}

Figure~\ref{fig:system} provides an overview of \method{}. Given the user query $q$ and the root URL $u_r$, \method{} first constructs and analyzes the website structure to identify a candidate set of URLs.
Under a limited navigation budget, \method{} models URL selection as a Multi-Armed Bandit (MAB) problem and applies Thompson sampling to adaptively prioritize promising URLs during navigation.
For each selected URL, the web navigation agent interacts with the browser to navigate starting from that URL.
After each navigation attempt, the reflection agent evaluates the navigation trajectory and provides feedback to update the bandit posterior for the next navigation attempt.

\subsection{Global Structure Analysis}
\label{sec: structure analysis}

Existing web navigation agents typically begin their exploration from the root URL~\cite{gur2024real, zhou2024language}. While effective for small-scale sites, this strategy becomes inefficient for large-scale websites with thousands of pages, as it forces the agent to incrementally traverse the enormous search space. To address this limitation, \method{} constructs and analyzes the global hierarchical structure of a website to identify a set of query-related URLs as starting points, denoted as $\mathcal{U}$, prior to navigation.

Specifically, given the root URL $u_r$, \method{} first performs lightweight web crawling via breadth-first search (BFS) to collect a set of reachable URLs starting from $u_r$. During crawling, \method{} filters out URLs that point to non-HTML files (e.g., images and videos) as well as URLs that lead to external websites, retaining only pages within the same domain. To ensure the process completes within a reasonable time, we set a limit $\tau$ on the maximum number of pages to crawl. To ensure simplicity and cost efficiency, \method{} employs the BM25 ranking algorithm~\cite{robertson2009probabilistic} to score these URLs based on the semantic similarity of their content to the query $q$. The top 10 URLs are added to $\mathcal{U}$. We describe the implementation details of this web crawling process in Appendix~\ref{sec: detail_web_crawling}.

However, some websites contain a vast number of webpages that cannot be fully reached through crawling alone due to time and depth constraints. For example, arXiv has over 2.9M pages of research papers.\footnote{\href{https://arxiv.org/stats/main}{https://arxiv.org/stats/main}} Consequently, crawling the entire site is practically infeasible. In such cases, relying solely on web crawling is insufficient. As Google has already indexed most pages within a website’s hierarchy, \method{} asks the LLM to recommend search keywords based on the user query (Prompt shown in Table~\ref{tab:generate-keywords}). Then, it leverages the \texttt{site} operator of Google Search to retrieve additional relevant pages and add the top 10 results into $\mathcal{U}$.

\subsection{URL Prioritization and Selection}
\label{sec: thompson sampling}
After the global structure analysis component identifies the candidate URLs set $\mathcal{U}$, \method{} needs to prioritize these URLs for navigation under a limited navigation budget. 
Inspired by \citet{chakrabarti2008mortal}, we model this task as a multi-armed bandit (MAB) problem with finite-lifetime arms. Unlike \citet{chakrabarti2008mortal}, which assigns a fixed lifetime to each arm (e.g., an arm $i$ is deactivated after being selected $L_i$ times), the deactivation of an arm in our method is dynamically determined by the reflection agent based on the navigation history. We formalize this variant of the MAB problem as follows:
\vspace{-0.15cm}
\begin{itemize}
\vspace{-0.15cm}
    \item \textbf{Arms}: The set of candidate URLs $\mathcal{U}$. Each URL $u_i$ maintains a dynamic status, being either \textit{Active} or \textit{Exhausted}.
    \vspace{-0.15cm}
    \item \textbf{Selection Strategy}: A probabilistic strategy based on Thompson Sampling that selects the most promising URL from the set of currently active arms, denoted as $\mathcal{U}_{act}$.
    \vspace{-0.15cm}
    \item \textbf{State Transitions}: A dual-update logic that modifies both the probability distribution and the active status of the arm based on navigation outcomes.
\end{itemize}
We initialize the Beta distribution parameters ($\alpha_u, \beta_u$) for each arm based on the relevance between the web content of a URL and the user query. Specifically, we reuse the BM25 relevance scores calculated by the global structure analysis component to initialize these two parameters. Let $\lambda_u$ be the BM25 score of the content of URL $u$, we normalize $\lambda_u$ to $\rho_u$:
\[
\rho_u = \frac{\lambda_u - \min_{u' \in \mathcal{U}_r}(\lambda_{u'})}{\max_{u' \in \mathcal{U}_r}(\lambda_{u'}) - \min_{u' \in \mathcal{U}_r}(\lambda_{u'}) + \epsilon}
\]
Here, $\epsilon$ is a small constant for numerical stability. Then, we initialize $\alpha_u, \beta_u$ as follows:
\[
\alpha_{u}^{(0)} = 1 + \kappa \cdot \rho_u, \quad \beta_{u}^{(0)} = 1 + \kappa \cdot (1 - \rho_u)
\]
Here, $\kappa$ is the weight parameter of the relevance.

At each step $t$, the agent selects a URL $u^*$ to visit by sampling a value $\theta_u$ from the posterior Beta distribution of each \textit{Active} arm and choosing the maximum:
\[
u^{*} = \arg\max_{u \in \mathcal{U}_{act}}
\left\{ \theta_u^{(t)} : \theta_u^{(t)} \sim \mathrm{Beta}(\alpha_u^{(t)}, \beta_u^{(t)}) \right\}
\]

After each navigation, the reflection agent outputs a status based on the navigation trajectory (detailed in Section~\ref{sec: reflection agent}). We assign a Bernoulli reward $r \in \{0,1\}$ for each type of status. Given the reward value $r^{(t)}$ of the status, \method{} updates the distribution parameters as follows:
\begin{equation}
\label{eq:beta_update}
\alpha_{u}^{(t+1)} = \alpha_{u}^{(t)} + r^{(t)}, \beta_{u}^{(t+1)} = \beta_{u}^{(t)} + (1 - r^{(t)})
\end{equation}
Furthermore, if the status indicates the navigation of the URL reached a dead end, \method{} will mark this URL as exhausted so it will not be considered in future URL selection and navigation.

\subsection{Web Navigation Agent}
\label{sec: web navigation agent}
The input to the web navigation agent consists of the user query $q$ and the selected URL $u^{*}$, which is chosen by Thompson Sampling. The web navigation agent is constrained with a navigation budget $b$ to limit the maximum actions to perform. If the URL has been previously visited, \method{} retrieves the navigation trajectory and reflection summary of the previous visit from the episodic memory and appends them to the input of the navigation agent. The navigation agent then decides which actions to perform on the browser environment and receives browser observations in response iteratively.

Specifically, \method{} treats the browser environment as a plug-in component. This architectural choice allows the navigation agent to interact with different web browsing environments and action spaces. 
To ensure a fair comparison with baselines, we align our browser environment settings with those used by leading methods on each benchmark. For example, when evaluating on WebVoyager~\cite{he-etal-2024-webvoyager}, we adopt the Playwright-based environment\footnote{\url{https://github.com/microsoft/playwright-mcp}} used by AgentOccam. Similarly, for the WebWalkerQA benchmark~\cite{wu-etal-2025-webwalker}, we utilize the Crawl4AI environment\footnote{\url{https://github.com/unclecode/crawl4ai}} employed by WebWalker. Adopting the same browsing environments as these SOTA methods makes it easier to run the experiments and establish a fair comparison with these web agents.

Finally, the web navigation agent may end in one of two states. First, it may have generated a response indicating completion of the task. However, there is a chance that the task is not yet fully completed, especially for tasks that require multiple actions, e.g., scraping multiple pieces of information from a webpage, filling in multiple text fields, etc. Second, the agent may exhaust its navigation budget $b$ before completing the task. This case is more challenging. The agent may be following a promising path but is forced to terminate early due to budget constraints, or it may be pursuing an unpromising path that would ultimately lead to a dead end. To carefully evaluate the status of the navigation attempt, the web navigation agent shares its navigation history with the reflection agent and hands off control to it.

\subsection{Reflection Agent}
\label{sec: reflection agent}

We introduce a reflection agent to further analyze the navigation trajectory and the final output of this navigation to determine the navigation attempt's status. If the web navigation agent indicates task completion, the reflection agent will assess whether the actions performed in the navigation trajectory and final output satisfy the user query. The prompt for this reflection is shown in Table~\ref{tab:reflection-success} in Appendix~\ref{sec:prompts}. 
If the answer is considered adequate by the reflection agent, \method{} terminates and outputs the final result. However, if the result is inadequate (e.g., the extracted information is partial or the interaction is incomplete), \method{} treats this as a promising path that requires further exploration. Consequently, it assigns a positive reward $r=1$ to update the Beta distribution parameters (Eq.~\ref{eq:beta_update}), thereby increasing the probability of re-selecting this URL in future iterations.

In the case of budget exhaustion, the reflection agent will evaluate whether the current navigation trajectory remains promising and should be continued. The prompt is shown in Table~\ref{tab:reflection-failure} in Appendix~\ref{sec:prompts}.
If the reflection agent deems the page is relevant but requires more navigation to complete the task, \method{} assigns a reward $r=1$ to maintain a high probability of selection. Otherwise, \method{} assigns a negative reward $r=0$, which updates the Beta distribution to discourage the agent from revisiting unpromising URLs.

The resulting status is then stored in the episodic memory and returned to the Thompson sampling selector to update the probability distribution. 
Furthermore, \method{} stores the navigation trajectory, the final output of this navigation, and the reflection in the episodic memory. In the next round of navigation, if the same URL is visited again, this information will be retrieved to help the web navigation agent to make a more informed decision without repeating the actions on the same URL.

\section{Experiment}
\subsection{Experimental Settings}
\label{sec: experimental_settings}
\paragraph{Benchmarks}
We conduct experiments on two web navigation benchmarks, WebVoyager~\cite{he-etal-2024-webvoyager} and WebWalkerQA~\cite{wu-etal-2025-webwalker}. WebVoyager consists of web navigation tasks on popular real-world websites, such as Amazon and Coursera. To evaluate \method{} on WebVoyager without the subjectivity introduced by human annotations, we follow AgentOccam~\cite{yang2025agentoccam} to use the 129 filtered QA tasks with golden answers. WebWalkerQA is a benchmark that contains 680 web navigation tasks across websites in four different domains, including conference, education, organization, and game. Furthermore, WebWalkerQA includes both single-source and multi-source QA tasks. Single-source tasks require deep exploration from the root URL to locate information on a single page, whereas multi-source tasks require integrating details from multiple distinct pages to answer a user query.

\paragraph{Comparison Baselines}
We compare \method{} with two prior methods: (1) AgentOccam~\cite{yang2025agentoccam}, a web agent that refines the action and observation space to better align with the LLM’s capabilities; and (2) WebWalker~\cite{wu-etal-2025-webwalker}, a web agent that adopts an explore–critic paradigm. These two baselines are the best-performing methods on the WebVoyager and WebWalkerQA benchmarks, respectively.

\paragraph{Base LLMs} We experiment with five different LLMs as the backbone of \method{}, including the GPT-5-mini model and Qwen3-\{4, 8, 14, 32\}B models. We access all models with their official API. For the GPT-5-mini model, we use the default parameter settings of the OpenAI API. For the Qwen3 models, we disable the thinking mode and set the temperature to 0.7 and top\_p to 0.8 following the official best-practice guideline.

\paragraph{Implementation Details}
In our experiments, we limit both the navigation budget $b$ and Thompson sampling iterations of \method{} to 10. We run all experiments in a single run.

\subsection{Results and Analysis}
\label{sec:results}
\begin{table}[htbp]
\centering
\small
\begin{tabular}{l|cccc|c}
\toprule
\multirow{2}{*}{Method} & \multicolumn{4}{c|}{\cellcolor{dustyblue}Qwen3} & \multicolumn{1}{c}{\cellcolor{sagegreen}GPT-5} \\
 & \cellcolor{dustyblue}4B & \cellcolor{dustyblue}8B & \cellcolor{dustyblue}14B & \cellcolor{dustyblue}32B & \cellcolor{sagegreen}mini \\
\midrule
WebWalker & 14.73 & 12.40 & 12.40 & 14.73 & 16.28 \\
AgentOccam & 22.48 & 20.93 & 25.58 & 34.11 & 56.25 \\
\method{} & \textbf{25.58} & \textbf{27.13} & \textbf{30.23} & \textbf{37.98} & \textbf{63.57} \\
\bottomrule
\end{tabular}
\caption{Comparison of the success rate (SR) of \method{} with baselines on WebVoyager.}
\label{tab:webvoyager_results}
\end{table}

\begin{table*}[t]
\centering
\small
\resizebox{0.99\textwidth}{!}{
\begin{tabular}{ll|cccc|cccc|c}
\toprule
& & \multicolumn{4}{c|}{{Single-source QA}} & \multicolumn{4}{c|}{{Multi-source QA}} & {Overall} \\
{Model} & {Method} & Easy & Medium & Hard & Overall & Easy & Medium & Hard & Overall & \\
\midrule
\headerblue \multicolumn{11}{c}{\textit{Open-Sourced LLMs}} \\
\midrule
\multirow{3}{*}{Qwen3-4B}
& WebWalker & 28.75 & 16.43 & 5.83 & 15.59 & 16.25 & 10.00 & 4.17 & 9.41 & 12.50 \\
& AgentOccam & 7.50 & 2.86 & 1.67 & 3.53 & 11.25 & 4.29 & 3.33 & 5.59 & 4.56 \\
& \method{} & 21.25 & 26.43 & 17.50 & 22.06 & 11.25 & 12.14 & 12.50 & 12.06 & \textbf{17.06} \\
\midrule
\multirow{3}{*}{Qwen3-8B}
& WebWalker & 15.00 & 18.57 & 13.33 & 15.88 & 17.50 & 8.57 & 5.00 & 9.41 & 12.65 \\
& AgentOccam & 8.75 & 4.29 & 1.67 & 4.41 & 10.00 & 5.00 & 2.50 & 5.29 & 4.85 \\
& \method{} & 26.25 & 31.43 & 25.83 & 28.24 & 16.25 & 15.71 & 15.00 & 15.59 & \textbf{21.91} \\
\midrule
\multirow{3}{*}{Qwen3-14B}
& WebWalker & 35.00 & 20.71 & 10.83 & 20.59 & 15.00 & 7.86 & 5.83 & 8.82 & 14.71 \\
& AgentOccam & 11.25 & 5.00 & 1.67 & 5.29 & 8.75 & 5.71 & 8.33 & 7.35 & 6.32 \\
& \method{} & 35.00 & 37.14 & 25.83 & 32.65 & 20.00 & 19.29 & 19.17 & 19.41 & \textbf{26.03} \\
\midrule
\multirow{3}{*}{Qwen3-32B}
& WebWalker & 35.00 & 22.14 & 14.17 & 22.35 & 13.75 & 15.00 & 5.00 & 11.18 & 16.76 \\
& AgentOccam & 11.25 & 10.00 & 5.00 & 8.53 & 16.25 & 11.43 & 11.67 & 12.65 & 10.59 \\
& \method{} & 41.25 & 40.71 & 25.00 & 35.29 & 25.00 & 22.14 & 18.33 & 21.47 & \textbf{28.38} \\
\midrule
\headergreen \multicolumn{11}{c}{\textit{Closed-Sourced LLMs}} \\
\midrule
\multirow{3}{*}{GPT-5-mini}
& WebWalker & 35.00 & 32.86 & 21.67 & 29.41 & 33.75 & 22.14 & 14.17 & 22.06 & 25.74 \\
& AgentOccam & 26.25 & 20.00 & 13.33 & 19.12 & 30.00 & 18.57 & 19.17 & 21.47 & 20.29 \\
& \method{} & 63.75 & 64.29 & 54.17 & 60.59 & 43.75 & 50.71 & 37.50 & 44.41 & \textbf{52.50} \\
\bottomrule
\end{tabular}
}
\caption{Comparison of the success rate (SR) of \method{} with baselines on WebWalkerQA.}
\label{tab:webwalker_results}
\end{table*}

\begin{table}[tbp]
\centering
\small
\resizebox{0.99\linewidth}{!}{
\begin{tabular}{l|cccc|c}
\toprule
\multirow{2}{*}{Method} & \multicolumn{4}{c|}{\cellcolor{dustyblue}Qwen3} & \multicolumn{1}{c}{\cellcolor{sagegreen}GPT-5} \\
 & \cellcolor{dustyblue}4B & \cellcolor{dustyblue}8B & \cellcolor{dustyblue}14B & \cellcolor{dustyblue}32B & \cellcolor{sagegreen}mini \\
\hline
\headercolorlong 
\multicolumn{6}{c}{WebVoyager}\\
WebWalker    & 9.11 & 7.00 & 7.44 & 9.32 & 7.38 \\
AgentOccam  & 5.34 & 6.48 & 4.55 & 6.07 & 9.46 \\
\method{}    & 6.21 & 7.83 & 6.26 & 5.35 & 14.18 \\
\hline
\headercolorlong 
\multicolumn{6}{c}{WebWalkerQA}\\
WebWalker    & 8.71 & 9.62 & 7.92 & 8.62 & 10.38 \\
AgentOccam  & 12.84 & 6.88 & 7.65 & 14.33 & 10.09 \\
\method{}    & 5.92 & 10.89 & 7.75 & 7.49 & 19.13 \\
\bottomrule
\end{tabular}
}
\caption{Action count comparison on WebVoyager and WebWalkerQA.}
\label{tab:action_counts}
\end{table}
In this section, we present a comprehensive analysis of \method{}'s performance and efficiency on the WebVoyager and WebWalkerQA benchmarks.

\paragraph{Success Rate}
Table~\ref{tab:webvoyager_results} shows the success rate of \method{} compared with WebWalker and AgentOccam on the WebVoyager benchmark. \method{} achieves the best performance across all settings. For instance, with GPT-5-mini as the backbone, \method{} achieves a 63.6\% success rate, surpassing AgentOccam with a 7.3\% absolute improvement and WebWalker with a 47.3\% absolute improvement.

Table~\ref{tab:webwalker_results} presents the results on the WebWalkerQA benchmark. \method{} consistently outperforms WebWalker and AgentOccam in terms of overall success rate across all backbone models. For example, with the GPT-5-mini backbone, \method{} achieves an overall success rate of 52.5\%, substantially outperforming WebWalker (25.7\%) and AgentOccam (20.3\%). Comparing the performance across different difficulty levels, we observe that \method{} exhibits better robustness than both baselines. While the success rate naturally declines as the task difficulty level increases, \method{} maintains a significant lead over baselines on the most challenging queries.

Furthermore, \method{} demonstrates robust capabilities across both single-source QA and multi-source QA tasks. On single-source QA tasks, \method{} achieves a 31.2\% absolute improvement over the best baseline when using GPT-5-mini. The performance gain is even more pronounced on the more challenging multi-source QA tasks---\method{} doubles the performance of both AgentOccam and WebWalker when using GPT-5-mini.

We also analyze the impact of the backbone model size using the Qwen3 family. As shown in Table~\ref{tab:webwalker_results}, the performance of \method{} scales monotonically with model size, improving from 17.1\% with Qwen3-4B to 28.4\% with Qwen3-32B.

\begin{figure}[htbp]
    \centering
    \begin{subfigure}[b]{0.43\textwidth}
        \centering
        \includegraphics[width=\linewidth]{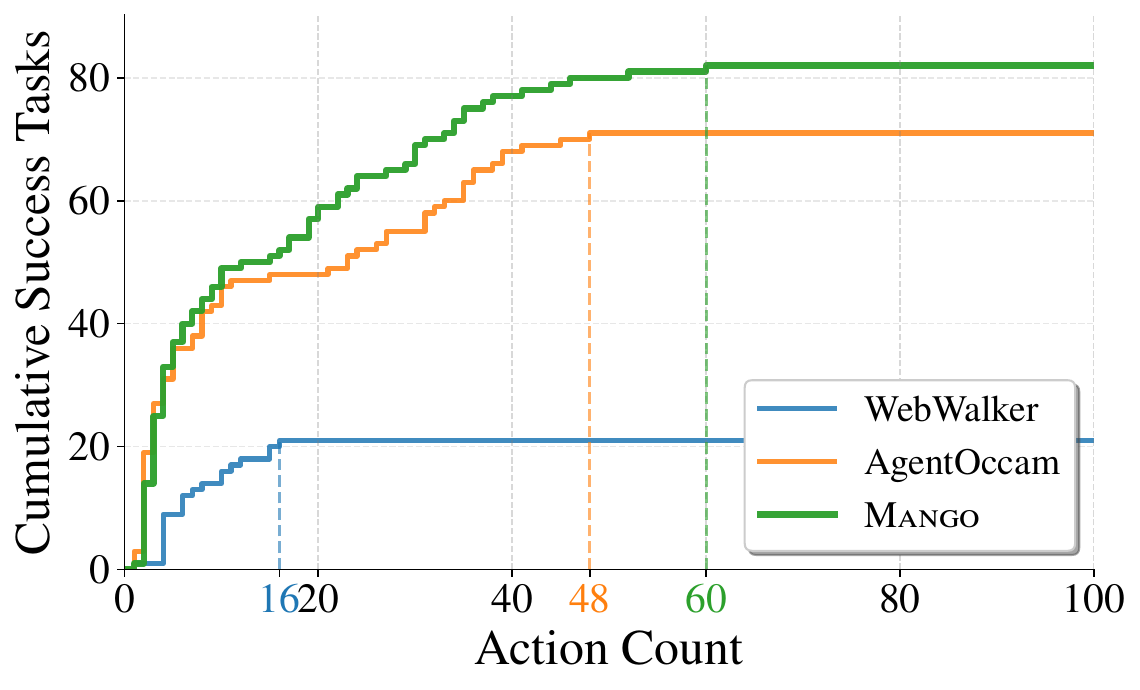}
        \caption{WebVoyager}
        \label{fig:cumulative_success_webvoyager}
    \end{subfigure}
    \hfill
    \begin{subfigure}[b]{0.43\textwidth}
        \centering
        \includegraphics[width=\linewidth]{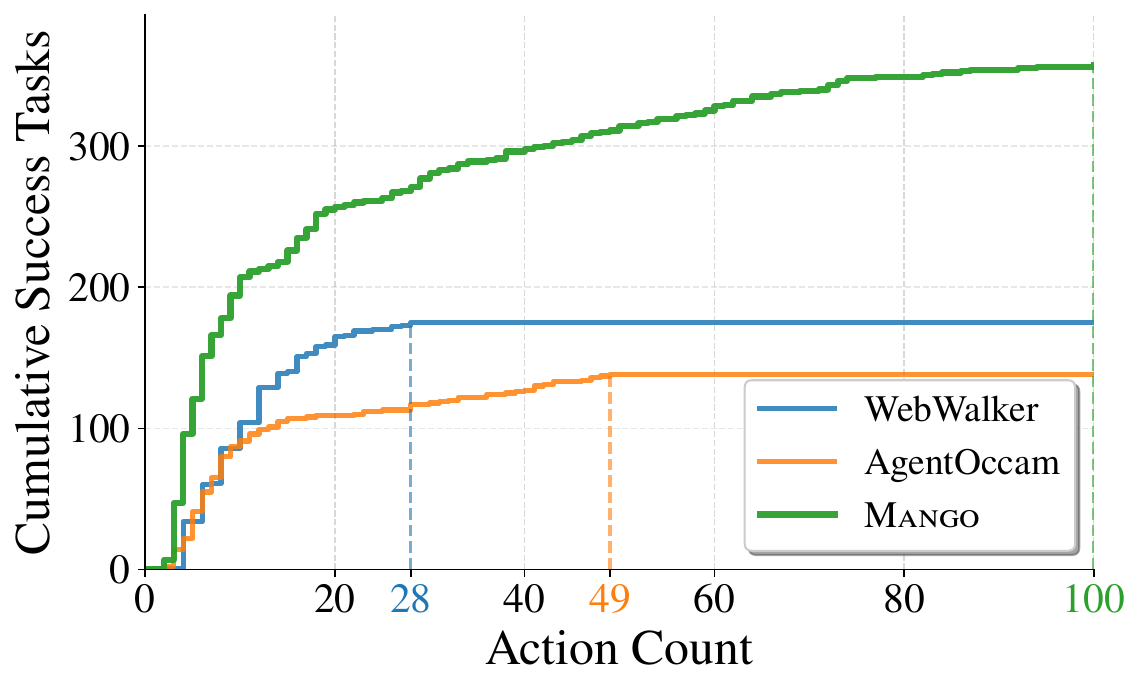}
        \caption{WebWalkerQA}
        \label{fig:cumulative_success_webwalker}
    \end{subfigure}
    \caption{Cumulative number of successful tasks relative to the action count using GPT-5-mini on WebVoyager (Figure~\ref{fig:cumulative_success_webvoyager}) and WebWalkerQA (Figure~\ref{fig:cumulative_success_webwalker}).}
    \label{fig:cumulative_success}
\end{figure}

\paragraph{Action Count} We count the total number of actions performed by \method{} and the baseline methods for each task to measure their efficiency. 
Table~\ref{tab:action_counts} reports the average action count of different methods on WebVoyager and WebWalkerQA. When using Qwen3 models, \method{} requires competitive or lower action counts while consistently obtaining higher success rates. Yet when using GPT-5-mini, \method{} takes more actions than both baselines on both benchmarks. An in-depth analysis reveals that this increase of actions is because \method{} solves more complex, long-horizon tasks when using GPT-5-mini. As illustrated in Figure~\ref{fig:cumulative_success}, WebWalker and AgentOccam plateau early on both benchmarks. Specifically, on WebWalkerQA (Figure~\ref{fig:cumulative_success_webwalker}), they fail to complete any more tasks after 28 and 49 action counts, respectively. In contrast, \method{} continues to accumulate successes well beyond 50 to 100 actions. This suggests that the higher average action count is driven by \method{}'s ability to persevere and succeed on challenging tasks that require more actions to complete.

Finally, we analyze the impact of model size on the number of actions within the Qwen3 family. As shown in Table~\ref{tab:action_counts}, the average action count remains relatively stable despite the significant improvement in success rates (e.g., from 17.1\% with Qwen3-4B to 28.4\% with Qwen3-32B on WebWalkerQA). This indicates that larger models in the Qwen3 family improve performance primarily through more accurate decision-making rather than by simply extending the exploration depth, thereby maintaining navigation efficiency even as their capability to handle complex tasks increases.

\subsection{Ablation Study}
\label{sec:ablation study}
\begin{table}[htbp]
\centering
\small
\resizebox{0.99\linewidth}{!}{
\begin{tabular}{l|cccc|c}
\toprule
\multirow{2}{*}{Method} & \multicolumn{4}{c|}{\cellcolor{dustyblue}Qwen3} & \multicolumn{1}{c}{\cellcolor{sagegreen}GPT-5} \\
 & \cellcolor{dustyblue}4B & \cellcolor{dustyblue}8B & \cellcolor{dustyblue}14B & \cellcolor{dustyblue}32B & \cellcolor{sagegreen}mini \\
\hline
\headercolorlong 
\multicolumn{6}{c}{WebVoyager}\\
\method{} $_{\text{random}}$ & 17.83 & 20.93 & 21.71 & 27.13 & 56.59 \\
\method{} $_{\text{google}}$ & 20.16 & 20.93 & 24.03 & 32.56 & 59.69 \\
\method{} $_{\text{MCTS}}$ & 18.60 & 19.38 & 20.93 & 23.26 & 46.51 \\
\method{} & \textbf{25.58} & \textbf{27.13} & \textbf{30.23} & \textbf{37.98} & \textbf{63.57} \\

\headercolorlong
\hline
\multicolumn{6}{c}{WebWalkerQA}\\
\method{} $_{\text{random}}$ & 10.15 & 13.97 & 16.47 & 19.85 & 47.50 \\
\method{} $_{\text{google}}$ & 15.59 & 18.38 & 23.09 & 25.88 & 49.41 \\
\method{} $_{\text{MCTS}}$ & 11.91 & 13.53 & 15.74 & 16.47 & 42.21 \\
\method{} & \textbf{17.06} & \textbf{21.91} & \textbf{26.03} & \textbf{28.38} & \textbf{52.50} \\
\bottomrule
\end{tabular}
}
\caption{Comparison of success rates (SR) between \method{} and three variants on WebVoyager and WebWalkerQA.}
\label{tab:ablation}
\end{table}

To validate the effectiveness of each component in \method{}, we conduct an ablation study by comparing \method{} with three variants. 

We first evaluate the impact of the global structure analysis component by modifying the candidate URL set generation strategy with two variants: \method{}$_{\text{random}}$ and \method{}$_{\text{google}}$.
\method{}$_{\text{random}}$ is a simple method that randomly selects a candidate URL set from the entire set of reachable pages collected during web crawling.
\method{}$_{\text{google}}$ constructs the candidate set $\mathcal{U}$ solely using URLs retrieved by Google Search.

Second, we assess the contribution of our URL prioritization and selection algorithm by comparing \method{} with an agentic search strategy, Monte Carlo Tree Search (MCTS), as introduced in Section~\ref{sec: agetic_memory}. We implement this variant inspired by both LATS~\cite{zhou2024language}, a general agent leveraging MCTS, and WebPilot~\cite{zhang2024webpilot}, a web agent leveraging a variant of MCTS. Since the implementation of WebPilot is not released publicly, we adopt the one-step simulation concept from it and retain the standard MCTS search for agents following LATS.
We connect the MCTS navigation agent after our Global Structure Analysis component, initializing the state using the candidate URL set $\mathcal{U}$ and performing MCTS to navigate the pages. We describe the implementation details in Appendix~\ref{sec: detail_mcts}.

As shown in Table~\ref{tab:ablation}, \method{} consistently outperforms all variants across both benchmarks. When comparing URL selection strategies, \method{}$_{\text{random}}$ exhibits lower performance (e.g., 56.59\% SR on WebVoyager with GPT-5-mini) compared to \method{}. While \method{}$_{\text{google}}$ performs better than the random baseline, it still lags behind the full method. This demonstrates that relying solely on Google Search is insufficient.

Furthermore, the comparison with \method{}$_{\text{MCTS}}$ reveals the superiority of our Thompson Sampling approach for budget-constrained web navigation. \method{} significantly outperforms \method{}$_{\text{MCTS}}$ across all settings. For instance, on WebVoyager with the GPT-5-mini backbone, \method{} achieves a success rate of 63.57\% compared to 46.51\% for \method{}$_{\text{MCTS}}$. This is primarily because MCTS requires a large number of interaction steps to expand the search tree and estimate state values, which becomes impractical under strict budget constraints. In contrast, Thompson Sampling can rapidly balance exploration and exploitation without simulation, allowing the agent to allocate its limited navigation budget more efficiently toward the most promising candidate URLs.

We further justify our design choices of the relevance scoring mechanism, the URL prioritization strategy, and the episodic memory component through additional ablation studies in Appendix~\ref{sec:additional_experiments}.

\subsection{Sensitivity Analysis}
\label{sec: sensitivity}
\begin{figure}[htbp]                                         
  \centering
  \begin{subfigure}[b]{0.23\textwidth}
      \centering
      \includegraphics[width=\linewidth]{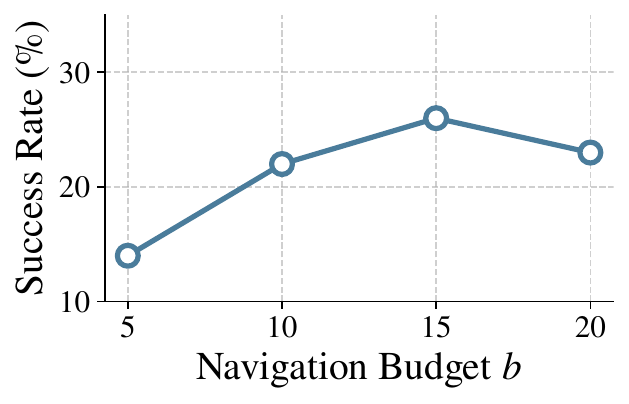}
      \caption{}                                                                                                
      \label{fig:navigation_budget}
  \end{subfigure}                                                                                               
  \hfill                                                   
  \begin{subfigure}[b]{0.23\textwidth}
      \centering                                                                                                
      \includegraphics[width=\linewidth]{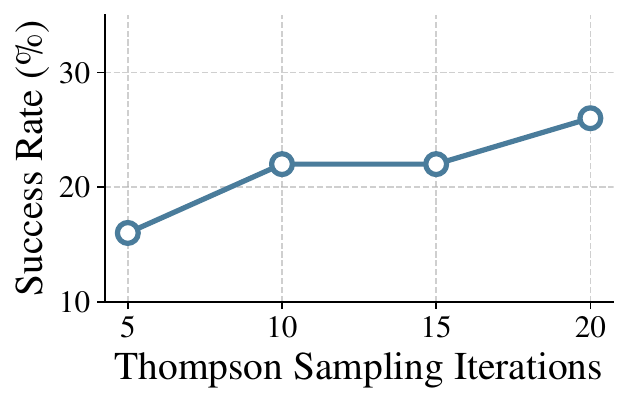}
      \caption{}                                                                                                
      \label{fig:ts_iteration}                             
  \end{subfigure}
  \\[0.5em]
  \begin{subfigure}[b]{0.23\textwidth}                                                                          
      \centering
      \includegraphics[width=\linewidth]{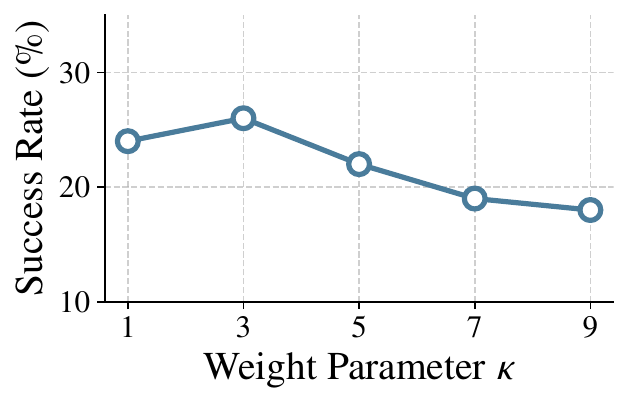}                                  
      \caption{}                                                                                                
      \label{fig:kappa}
  \end{subfigure}                                                                                               
  \hfill                                                   
  \begin{subfigure}[b]{0.23\textwidth}
      \centering                                                                                                
      \includegraphics[width=\linewidth]{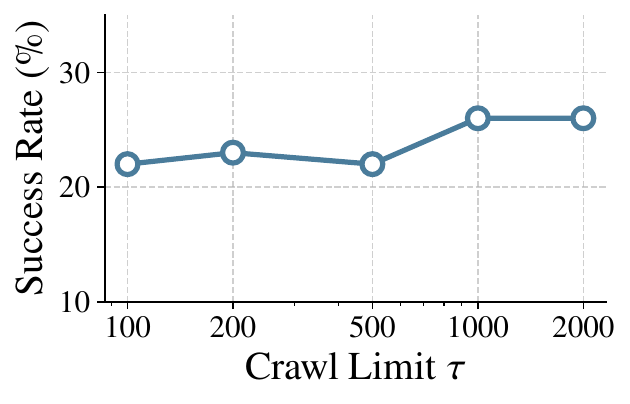}
      \caption{}                                                                                                
      \label{fig:crawl_limit}                              
  \end{subfigure}
  \hfill                                                                                                        
  \begin{subfigure}[b]{0.23\textwidth}
      \centering                                                                                                
      \includegraphics[width=\linewidth]{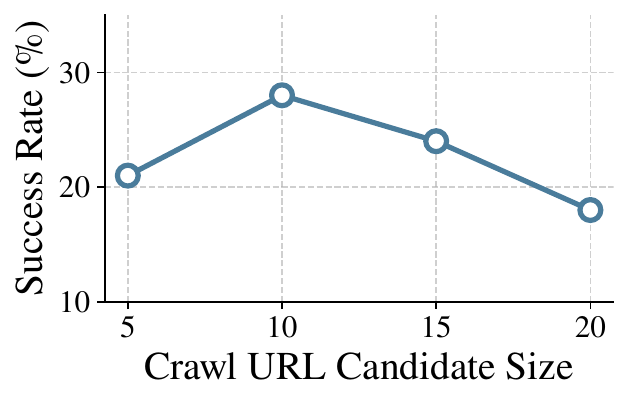}
      \caption{}
      \label{fig:crawl_url_size}                                                                                
  \end{subfigure}
  \hfill                                                                                                        
  \begin{subfigure}[b]{0.23\textwidth}                     
      \centering
      \includegraphics[width=\linewidth]{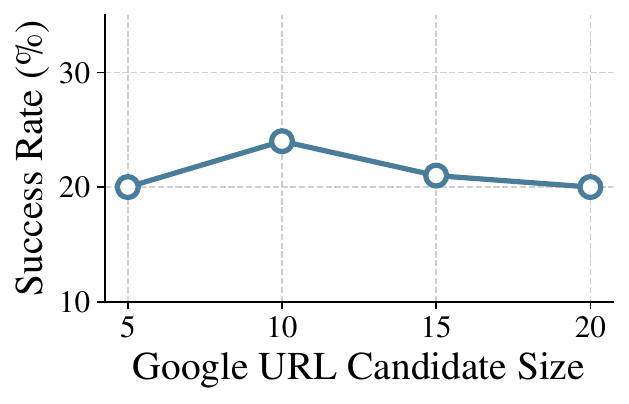}
      \caption{}                                                                                                
      \label{fig:google_url_size}
  \end{subfigure}                                                                                               
  \caption{Sensitivity analysis of \method{} regarding key hyperparameters: the navigation budget per URL selection $b$, the number of Thompson Sampling iterations, the weight parameter $\kappa$, the crawl limit $\tau$, and the candidate set sizes of Crawling and Google Search.}
  \label{fig:parameter}                                                                                         
\end{figure}
We analyze the sensitivity of \method{} to five key hyperparameters: the navigation budget per URL selection $b$, the number of Thompson Sampling iterations, the weight parameter $\kappa$, the crawl limit $\tau$, and the candidate set size. We run \method{} in different budget settings on 100 tasks from WebWalkerQA using Qwen3-32B.

\paragraph{Navigation Budget $b$.}
As shown in Figure~\ref{fig:navigation_budget}, the success rate improves as $b$ increases from 5 to 15. However, performance declines slightly when $b$ is further increased to 20, suggesting that excessive actions on a single navigation attempt may cause the agent to confabulate or get lost in irrelevant paths.

\paragraph{Thompson Sampling Iterations.}
Figure~\ref{fig:ts_iteration} shows a positive correlation between the number of Thompson Sampling iterations and the success rate. This confirms that allocating more iterations allows the bandit to better balance exploration and exploitation across candidate URLs, leading to more effective navigation.

\paragraph{Weight Parameter $\kappa$.}
The parameter $\kappa$ controls the strength of the initial BM25 relevance score when initializing the Beta distribution for Thompson Sampling. As shown in Figure~\ref{fig:kappa}, the success rate peaks at 26\% when $\kappa{=}3$. As $\kappa$ increases to 7 and 9, performance drops to 19\% and 18\%, respectively. A high $\kappa$ makes the initial prior too rigid, overpowering the reward signal from the reflection agent and preventing the bandit from dynamically adapting to navigation outcomes.

\paragraph{Crawl Limit $\tau$.}
Figure~\ref{fig:crawl_limit} shows that performance steadily improves as $\tau$ increases from 100 (22\% SR) to 1000 (26\% SR). However, doubling $\tau$ to 2000 yields no further improvement. Therefore, $\tau{=}1000$ serves as an optimal sweet spot, maximizing structural coverage without incurring the exponential time and computational costs associated with deeper crawling.

\paragraph{Candidate Set Size.}
Figures~\ref{fig:crawl_url_size} and~\ref{fig:google_url_size} show the effect of varying the number of candidates drawn from crawled URLs and Google search results, respectively. For both sources, Top-10 yields the best performance (28\% and 24\% SR). Since the navigation budget and Thompson Sampling iterations are capped at 10, an overly large combined candidate pool spreads the exploration budget too thin. Taking Top-10 from both sources thus strikes the ideal trade-off.

\subsection{Failure Case Analysis}

To understand the limitations of \method{}, we manually inspected 323 failure cases from the WebWalkerQA benchmark using GPT-5-mini as the backbone model. We identified five distinct error patterns:
\vspace{-0.15cm}
\begin{itemize}
    \item \textbf{Exceed Budget (52.4\%):} This is the most common failure mode, where the agent exhausts the navigation budget before locating the target information. We identify two possible reasons for this issue. First, \method{} relies on lightweight crawling and search-based augmentation to construct a global view, which may provide incomplete coverage for very large websites with thousands of webpages. If the target information is buried deep in such websites, \method{} may not be able to find it with a limited budget. Second, the bandit-based URL selection depends on the quality of the initial candidate set, and errors in relevance estimation can result in suboptimal early choices that incur irreversible budget costs under strict action limits.
    \vspace{-0.15cm}
    \item \textbf{Locating Wrongly (24.6\%):} This error occurs when the agent navigates to an incorrect or irrelevant sub-page. \method{} may be misled by ambiguous links, resulting in navigation trajectories that deviate from the target information source.
    \vspace{-0.15cm}
    \item \textbf{Reasoning Error (15.4\%):} In these cases, the agent successfully navigates to the correct page containing the target information but fails to generate the correct answer. For example, Task 140 asks for the student ID under a specific condition from the class of 2020, but \method{} outputs the student ID that satisfies the corresponding condition for the class of 2022. This indicates a failure in the underlying LLM's reading comprehension or reasoning capabilities, leading to hallucinated or incorrect extraction of details from the correct context.
    \vspace{-0.15cm}
    \item \textbf{Out-of-date Golden Answers (5.6\%):} In a small portion of tasks in WebWalkerQA, the agent successfully retrieves the correct answer to the user query, but the answer is judged as incorrect because the golden answer is out-of-date. For example, Task 155 in WebWalkerQA asks how many fiscal years in a row Sony fulfilled the goal of using renewable electricity in China. \method{} navigates to the correct news page published in 2025 and answers ``5 years'' (2020-2024). However, since this benchmark was created in 2024, which was one year earlier than 2025, the golden answer was labeled as ``4 years''.
    \vspace{-0.15cm}
    \item \textbf{Reflection Error (2.0\%):} The rarest error type involves the reflection agent incorrectly assessing the navigation state. Here, the reflection module prematurely classifies a partial or incorrect response as ``adequate,'' causing the agent to terminate the navigation session before the user query is fully satisfied.
\end{itemize}

\section{Conclusion}
We propose \method{}, a web navigation framework that leverages global website structure to improve navigation efficiency under limited budgets. By identifying query-relevant entry points and modeling URL selection as a multi-armed bandit problem with Thompson Sampling to prioritize the URL, \method{} allocates exploration efforts more effectively. Experiments on WebVoyager and WebWalkerQA show that \method{} consistently outperforms best baselines across five backbone models.

\section{Limitation}

\method{} constructs a global view using lightweight crawling and search-based augmentation, which does not guarantee full coverage for large, dynamic, or deeply nested websites. In practice, many real-world websites (e.g., e-commerce platforms like Amazon) contain hundreds of thousands of pages, making exhaustive crawling both time-prohibitive and unnecessary. Therefore, \method{} deliberately constructs only a partial approximation of the site structure, focusing on representative and high-relevance pages. As a result, tasks with target information buried far down the hierarchy may still exceed the navigation budget. This indicates that partial structural visibility remains a bottleneck for long-horizon web navigation.

The bandit-based URL selection in \method{} depends on the quality of the initial candidate set generated during global structure analysis. Errors in relevance estimation or keyword generation can introduce suboptimal entry points, leading to early budget misallocation. While Thompson Sampling mitigates this over time, incorrect early decisions are costly under strict action limits.

Even when navigation succeeds, \method{} can fail due to reasoning errors, such as incorrect extraction or hallucinated details. These failures are orthogonal to navigation quality and cannot be addressed solely through improved exploration strategies. This limitation is shared by most current LLM-based web agents.

\method{} often achieves higher success rates by continuing exploration beyond where baseline agents plateau, which can increase action counts. While beneficial for solving long-horizon tasks, this behavior may be undesirable in latency-sensitive or cost-sensitive settings.

\section*{Acknowledgments}
We sincerely thank the anonymous reviewers for their constructive feedback. This work was supported by NSF Grant ITE-2333736.

\bibliography{references}

\appendix
\appendix

\section{Implementation Details}
\label{sec: implementation details}
\subsection{Web Crawling}
\label{sec: detail_web_crawling}
In this section, we describe our implementation details for the light-weight web crawling used in the global structure analysis component ( Section~\ref{sec: structure analysis}).
We implement the web crawling component using the open-source tool Crawl4AI~\footnote{\url{https://github.com/unclecode/crawl4ai}}. We employ the \texttt{BFSDeepCrawlStrategy} to traverse the website structure and extract page content in markdown format. To ensure the process completes within a reasonable timeframe, we set the \texttt{max\_pages} (i.e., $\tau$ in Section~\ref{sec: structure analysis}) parameter to 1000. Additionally, we set \texttt{include\_external} to \texttt{False} and \texttt{exclude\_all\_images} to \texttt{True}. These constraints ensure the crawler remains focused on the target domain's hierarchy and filters out non-navigational assets such as image paths.

\subsection{\method{}$_{\text{MCTS}}$}
\label{sec: detail_mcts}
Following LATS~\cite{zhou2024lats}, \method{}$_{\text{MCTS}}$ has the same six operations: selection, expansion, evaluation, simulation, backpropagation, and reflection. %
To avoid exceeding the context length of LLMs due to long page content, instead of expanding the currently selected node until a terminal state is reached as in LATS, we leverage the one-step simulation strategy from WebPilot. Specifically, we perform the action that has the highest score from the evaluation step and ask the LLM to generate a reflection based on the observation after performing this action. We set the maximum number of trajectories to sample, the number of times to prompt during expansion, and the number of times to prompt for state evaluation all to 10. As we set both the navigation budget $b$ and the number of Thompson sampling iterations of \method{} to 10 (Section~\ref{sec: experimental_settings}), we limit the maximum number of actions to 100 to ensure a fair comparison.

\section{Additional Experiments and Analyses}
\label{sec:additional_experiments}
In this section, we present additional experiments and analyses to justify our design choices regarding relevance scoring, URL prioritization, and the episodic memory component. We conduct experiments on the same subset of the WebWalkerQA benchmark as Section~\ref{sec: sensitivity}, using GPT-5-mini as the backbone model.
\begin{table*}[t]
\centering
\small
\begin{tabular}{l|ccc|ccc|c}
\toprule
& \multicolumn{3}{c|}{Single-source QA} & \multicolumn{3}{c|}{Multi-source QA} & Overall \\
Method & Easy & Medium & Hard & Easy & Medium & Hard & \\
\midrule
\textsc{Mango}$_{\text{Embedding}}$ & 50.0 & 77.8 & 53.3 & 75.0 & 44.0 & 44.4 & 53.0 \\
\textsc{Mango}$_{\text{Greedy}}$ & 25.0 & 72.2 & 66.7 & 75.0 & 40.0 & 44.4 & 51.0 \\
\textsc{Mango}$_{\text{No\_Mem}}$ & 25.0 & 61.1 & 73.3 & 50.0 & 36.0 & 44.4 & 47.0 \\
\method{} & 50.0 & 72.2 & 60.0 & 75.0 & 48.0 & 44.4 & \textbf{55.0} \\
\bottomrule
\end{tabular}
\caption{Additional ablation studies on \method{}, evaluated on a subset of WebWalkerQA with GPT-5-mini as the backbone. We compare \method{} with variants that (1) replace BM25 with semantic embeddings, (2) replace Thompson Sampling with a greedy ranking strategy, and (3) remove the episodic memory component.}
\label{tab:additional_ablation}
\end{table*}

\subsection{Impact of Relevance Scoring Mechanisms}
\label{sec:relevance}
In the URL Prioritization and Selection component (Section~\ref{sec: thompson sampling}), \method{} employs BM25 to score candidate URLs based on their relevance to the user query. While semantic embedding models could potentially capture richer contextual signals, computing embeddings for up to 1,000 crawled pages per task incurs substantial computational overhead. To validate our choice of BM25, we compare \method{} against a variant, $\text{\method{}}_{\text{Embedding}}$, which replaces BM25 with semantic similarity scores computed by the \texttt{KaLM-Embedding-V2.5} model\footnote{\url{https://huggingface.co/KaLM-Embedding/KaLM-embedding-multilingual-mini-instruct-v2.5}}.
As shown in Table~\ref{tab:additional_ablation}, \method{} achieves a 55.0\% overall success rate, slightly outperforming the embedding-based variant at 53.0\%. We found that using a pre-trained embedding model sometimes struggled to distinguish page URLs that have subtle differences in wording, while BM25 is more sensitive to word differences. Nevertheless, fine-tuning an embedding model specifically for this task or performing semantic analysis on the page content instead of just the page URL would presumably lead to better performance, but would also lead to higher development cost or runtime cost.

\subsection{Effectiveness of Thompson Sampling over Greedy Selection}
\label{sec:mab_greedy}
To demonstrate that our MAB formulation provides meaningful benefits over simpler ranking approaches, we compare \method{} with a greedy variant, $\text{\method{}}_{\text{Greedy}}$, which ranks candidate URLs by their initial BM25 scores and visits them in descending order without dynamic updates from the reflection agent.
As shown in Table~\ref{tab:additional_ablation}, \method{} outperforms $\text{\method{}}_{\text{Greedy}}$ with a 4.0\% absolute improvement in overall success rate. This confirms that Thompson Sampling effectively leverages feedback from the reflection agent to dynamically update the posterior distribution, enabling \method{} to adaptively prune unpromising URLs and reallocate the navigation budget toward more promising candidates.

\subsection{Ablation on Episodic Memory}
\label{sec:memory_ablation}
The episodic memory component in \method{} stores navigation trajectories and reflections, which are retrieved to inform subsequent navigation attempts on the same URL (Section~\ref{sec: reflection agent}). To isolate the contribution of this component, we evaluate a variant, $\text{\method{}}_{\text{No\_Memory}}$, in which the episodic memory module is removed.
As shown in Table~\ref{tab:additional_ablation}, removing the episodic memory causes the overall success rate to drop from 55.0\% to 47.0\%, an 8.0\% absolute decrease. This is consistent with our design motivation: without access to prior trajectories and reflections, the navigation agent is prone to repeating the same actions when revisiting a URL, leading to redundant exploration and premature budget exhaustion.

\section{Prompts}
\label{sec:prompts}
In this section, we present the full prompts of \method{}. Table~\ref{tab:generate-keywords} shows the prompt of generating keywords described in Section~\ref{sec: structure analysis}. Table~\ref{tab:simple-navigation-agent} shows the prompt of the navigation agent (Section~\ref{sec: web navigation agent}). Table~\ref{tab:reflection-success} and \ref{tab:reflection-failure} shows the prompts of the reflection agent (Section~\ref{sec: reflection agent}).

\begin{table}[ht]
    \centering
    \small
    \begin{tabular}{>{\raggedright\arraybackslash\ttfamily}p{0.98\linewidth}<{}}
        \toprule
            You are a search expert.\\\\
            Transform the user's intent into a concise search query composed of space-separated keywords.\\
            Output only the final query.\\
        \bottomrule
    \end{tabular}
    \caption{
        Prompt for the Search Query Generator. It transforms complex user intents into concise keyword-based queries.
    }
    \label{tab:generate-keywords}
\end{table}
\begin{table}[ht]
      \centering
      \small
      \begin{tabular}{>{\raggedright\arraybackslash\ttfamily}p{0.98\linewidth}<{}}
          \toprule
              You are a Web Navigation Agent.\\\\

              You can call functions to visit websites as needed.\\
              You may also be provided with previous navigation history, including function calls, previous outputs, and reflections, to help inform your decisions.\\
              You may choose to continue navigating by revisiting a previously visited URL or by starting fresh from the root URL.\\

              \headercolorlong
              \textbf{TASK INSTRUCTIONS}\\
              1. Use the browser functions to visit and explore the target URL.\\
              2. Read the page content thoroughly.\\
              3. If you find new content that can answer the user query, generate an answer based on the page content. Otherwise, continue navigating to find relevant information.\\\\

              \headercolorlong
              \textbf{HANDOFF INSTRUCTIONS}\\
              Instead of outputting text, you must hand off control to the appropriate reflection agent based on your findings.\\\\

              \textbf{Case 1: Relevant Information Found}\\
              If you find new content that clearly answers the user query:\\
              \hspace{0.5cm}- Hand off to the \texttt{success\_reflection\_agent}.\\
              \hspace{0.5cm}- Pass \texttt{result} and \texttt{source} (the specific URL) to the handoff function.\\\\

              \textbf{Case 2: Stuck / Information Not Found}\\
              If you cannot find relevant information, reach a dead end, determine that the page content is entirely irrelevant, or cannot find new content after thorough exploration:\\
              \hspace{0.5cm}- Hand off to the \texttt{failure\_reflection\_agent}.\\
              \hspace{0.5cm}- You do not need to provide content, but ensure that you have explored the page sufficiently.\\\\

              \headercolorlong
              \textbf{TASK}\\
              User Query: \{USER\_QUERY\}\\
              Root URL: \{ROOT\_URL\}\\
          \bottomrule
      \end{tabular}
      \caption{
          Prompt for the Web Navigation Agent. This instruction guides the agent to explore pages using previous navigation history and hand off control to specific reflection agents based on success or failure.
      }
      \label{tab:simple-navigation-agent}
  \end{table}

\begin{table}[ht]
      \centering
      \small
      \begin{tabular}{>{\raggedright\arraybackslash\ttfamily}p{0.98\linewidth}<{}}
          \toprule
              You are a Navigation Decision Evaluator.\\\\

              You are reviewing a navigation session in which the agent has generated a response indicating task completion. Your goal is to determine whether the navigation actions and output fully satisfy the user's query.\\\\

              \headercolorlong
              \textbf{DECISION FRAMEWORK}\\
              A. adequate\\
              \hspace{0.5cm}- The final output and navigation trajectory provide a complete and comprehensive answer to the User Query. The answer needs to cover all questions of the User Query.\\
              \hspace{0.5cm}- No further navigation is needed.\\\\

              B. inadequate\\
              \hspace{0.5cm}- The output is partial or relevant, but does not fully answer the User Query.\\
              \hspace{0.5cm}- Further navigation on following links is likely to provide the missing information.\\\\

              \headercolorlong
              \textbf{OUTPUT FORMAT}\\
              Output a JSON object.\\
              \{\\
              \hspace{0.5cm}"status": {\color{blue} "adequate" | "inadequate"},\\
              \hspace{0.5cm}"reason": {\color{blue} "Explain why the current output is sufficient or why we should continue."},\\
              \hspace{0.5cm}"output": {\color{blue} "The response generated by the navigation agent."},\\
              \hspace{0.5cm}"source": {\color{blue} "The URL where the content was extracted from."}\\
              \}\\
          \bottomrule
      \end{tabular}
      \caption{
              Prompt for the Reflection Agent (Task Completion Case). This prompt is triggered when the navigation agent indicates task completion to determine if the agent should stop or continue.
      }
      \label{tab:reflection-success}
  \end{table}
\begin{table}[ht]
      \centering
      \small
      \begin{tabular}{>{\raggedright\arraybackslash\ttfamily}p{0.98\linewidth}<{}}
          \toprule
              You are a Navigation Decision Evaluator.\\\\

              The web navigation agent has exhausted its navigation budget before completing the task. Your goal is to analyze the navigation trajectory and the final URL to decide if this path is promising and should be continued, or if it should be abandoned.\\\\

              \headercolorlong
              \textbf{DECISION FRAMEWORK}\\
              A. feasible\\
              \hspace{0.5cm}- The answer was not found yet, but the current page is relevant to the User Query.\\
              \hspace{0.5cm}- The stop might be due to budget constraints, but the agent simply needs to visit more links or navigate deeper on this site.\\
              \hspace{0.5cm}- We should NOT give up on this path yet.\\\\

              B. infeasible\\
              \hspace{0.5cm}- The page is irrelevant, a dead end, or the site is broken.\\
              \hspace{0.5cm}- Repeated actions in the trajectory suggest no answer exists here.\\ 
              \hspace{0.5cm}- We should abandon this path.\\\\

              \headercolorlong
              \textbf{OUTPUT FORMAT}\\
              Output a JSON object.\\\\

              \{\\
              \hspace{0.5cm}"status": {\color{blue} "feasible" | "infeasible"},\\
              \hspace{0.5cm}"reason": {\color{blue} "Explain why the current trajectory is promising vs. why it leads to a dead end."}\\
              \}\\
          \bottomrule
      \end{tabular}
      \caption{
          Prompt for the Reflection Agent (Budget Exhaustion Case). This prompt is triggered when the agent exhausts its navigation budget, helping determine if the current trajectory remains promising.
      }
      \label{tab:reflection-failure}
  \end{table}

\section{Full Results}
\begin{table*}[t]
\centering
\small
\resizebox{0.99\textwidth}{!}{
\begin{tabular}{l|ccc|ccc|ccc|ccc|ccc}
\toprule
{Website} & \multicolumn{3}{c|}{\cellcolor{dustyblue}Qwen3-4B} & \multicolumn{3}{c|}{\cellcolor{dustyblue}Qwen3-8B} & \multicolumn{3}{c|}{\cellcolor{dustyblue}Qwen3-14B} & \multicolumn{3}{c|}{\cellcolor{dustyblue}Qwen3-32B} & \multicolumn{3}{c}{\cellcolor{sagegreen}{GPT-5-mini}} \\
 & \cellcolor{dustyblue} WW & \cellcolor{dustyblue}AO & \cellcolor{dustyblue}M & \cellcolor{dustyblue}WW & \cellcolor{dustyblue}AO & \cellcolor{dustyblue}M & \cellcolor{dustyblue}WW & \cellcolor{dustyblue}AO & \cellcolor{dustyblue}M & \cellcolor{dustyblue}WW & \cellcolor{dustyblue}AO & \cellcolor{dustyblue}M & \cellcolor{sagegreen}WW & \cellcolor{sagegreen}AO & \cellcolor{sagegreen}M \\
\midrule
Allrecipes & 25.0 & 0.0 & 100.0 & 50.0 & 25.0 & 25.0 & 25.0 & 0.0 & 50.0 & 25.0 & 0.0 & 75.0 & 25.0 & 75.0 & 75.0 \\
Amazon & 0.0 & 0.0 & 0.0 & 0.0 & 0.0 & 0.0 & 0.0 & 0.0 & 0.0 & 0.0 & 0.0 & 0.0 & 0.0 & 100.0 & 0.0 \\
Apple & 14.3 & 0.0 & 28.6 & 14.3 & 14.3 & 14.3 & 28.6 & 14.3 & 14.3 & 28.6 & 28.6 & 28.6 & 28.6 & 57.1 & 57.1 \\
ArXiv & 18.8 & 18.8 & 12.5 & 18.8 & 25.0 & 18.8 & 18.8 & 25.0 & 25.0 & 12.5 & 37.5 & 37.5 & 12.5 & 53.3 & 56.2 \\
BBC News & 0.0 & 0.0 & 50.0 & 0.0 & 0.0 & 50.0 & 0.0 & 0.0 & 50.0 & 0.0 & 50.0 & 50.0 & 0.0 & 50.0 & 50.0 \\
Booking & 0.0 & 50.0 & 0.0 & 0.0 & 50.0 & 0.0 & 0.0 & 50.0 & 50.0 & 0.0 & 50.0 & 50.0 & 0.0 & 100.0 & 100.0 \\
Cambridge Dictionary & 22.2 & 0.0 & 44.4 & 22.2 & 0.0 & 55.6 & 22.2 & 0.0 & 66.7 & 22.2 & 0.0 & 88.9 & 44.4 & 11.1 & 88.9 \\
Coursera & 50.0 & 0.0 & 100.0 & 0.0 & 0.0 & 50.0 & 50.0 & 0.0 & 50.0 & 50.0 & 0.0 & 50.0 & 50.0 & 50.0 & 100.0 \\
ESPN & 30.0 & 20.0 & 30.0 & 30.0 & 10.0 & 40.0 & 40.0 & 30.0 & 50.0 & 20.0 & 40.0 & 50.0 & 40.0 & 50.0 & 70.0 \\
Google Map & 0.0 & 33.3 & 0.0 & 0.0 & 11.1 & 0.0 & 0.0 & 22.2 & 0.0 & 0.0 & 55.6 & 33.3 & 0.0 & 55.6 & 77.8 \\
Google Search & 0.0 & 0.0 & 37.5 & 0.0 & 6.2 & 50.0 & 0.0 & 12.5 & 50.0 & 6.2 & 12.5 & 50.0 & 0.0 & 56.2 & 68.8 \\
Huggingface & 11.8 & 23.5 & 17.6 & 5.9 & 5.9 & 23.5 & 11.8 & 17.6 & 11.8 & 17.6 & 17.6 & 11.8 & 17.6 & 41.2 & 41.2 \\
Wolfram Alpha & 17.6 & 47.1 & 17.6 & 11.8 & 47.1 & 20.6 & 2.9 & 50.0 & 23.5 & 14.7 & 58.8 & 26.5 & 11.8 & 73.5 & 61.8 \\
\midrule
Overall & 14.7 & 22.5 & \textbf{25.6} & 12.4 & 20.9 & \textbf{27.1} & 12.4 & 25.6 & \textbf{30.2} & 14.7 & 34.1 & \textbf{38.0} & 16.3 & 56.2 & \textbf{63.6} \\
\bottomrule
\end{tabular}
}
\caption{The full success rate (SR) results categorized by websites on WebVoyager. WW refers to WebWalker, AO refers to AgentOccam, and M refers to \method{}.}
\label{tab:webvoyager_results_full}
\end{table*}
\begin{table*}[t]
\centering
\small
\resizebox{0.99\textwidth}{!}{
\begin{tabular}{l|cccc|cccc|cccc|cccc|cccc}
\toprule
{Website} & \multicolumn{4}{c|}{\cellcolor{dustyblue}\textbf{Qwen3-4B}} & \multicolumn{4}{c|}{\cellcolor{dustyblue}\textbf{Qwen3-8B}} & \multicolumn{4}{c|}{\cellcolor{dustyblue}\textbf{Qwen3-14B}} & \multicolumn{4}{c|}{\cellcolor{dustyblue}\textbf{Qwen3-32B}} & \multicolumn{4}{c}{\cellcolor{sagegreen}\textbf{GPT-5-mini}} \\
 & \cellcolor{dustyblue}R & \cellcolor{dustyblue}G & \cellcolor{dustyblue}MCTS & \cellcolor{dustyblue}M & \cellcolor{dustyblue}R & \cellcolor{dustyblue}G & \cellcolor{dustyblue}MCTS & \cellcolor{dustyblue}M & \cellcolor{dustyblue}R & \cellcolor{dustyblue}G & \cellcolor{dustyblue}MCTS & \cellcolor{dustyblue}M & \cellcolor{dustyblue}R & \cellcolor{dustyblue}G & \cellcolor{dustyblue}MCTS & \cellcolor{dustyblue}M & \cellcolor{sagegreen}R & \cellcolor{sagegreen}G & \cellcolor{sagegreen}MCTS & \cellcolor{sagegreen}M \\
\midrule
Allrecipes & 25.0 & 25.0 & 25.0 & 100.0 & 25.0 & 0.0 & 25.0 & 25.0 & 50.0 & 50.0 & 25.0 & 50.0 & 50.0 & 50.0 & 50.0 & 75.0 & 75.0 & 50.0 & 50.0 & 75.0 \\
Amazon & 0.0 & 0.0 & 0.0 & 0.0 & 0.0 & 0.0 & 0.0 & 0.0 & 0.0 & 0.0 & 0.0 & 0.0 & 100.0 & 0.0 & 0.0 & 0.0 & 0.0 & 0.0 & 0.0 & 0.0 \\
Apple & 0.0 & 0.0 & 14.3 & 28.6 & 14.3 & 14.3 & 14.3 & 14.3 & 14.3 & 0.0 & 28.6 & 14.3 & 14.3 & 28.6 & 28.6 & 28.6 & 28.6 & 42.9 & 28.6 & 57.1 \\
ArXiv & 31.2 & 12.5 & 18.8 & 12.5 & 18.8 & 25.0 & 18.8 & 18.8 & 31.2 & 31.2 & 18.8 & 25.0 & 31.2 & 25.0 & 18.8 & 37.5 & 68.8 & 62.5 & 37.5 & 56.2 \\
BBC News & 50.0 & 50.0 & 0.0 & 50.0 & 0.0 & 0.0 & 0.0 & 50.0 & 0.0 & 0.0 & 0.0 & 50.0 & 0.0 & 50.0 & 0.0 & 50.0 & 100.0 & 50.0 & 0.0 & 50.0 \\
Booking & 0.0 & 0.0 & 50.0 & 0.0 & 50.0 & 50.0 & 50.0 & 0.0 & 0.0 & 0.0 & 50.0 & 50.0 & 0.0 & 50.0 & 50.0 & 50.0 & 100.0 & 100.0 & 50.0 & 100.0 \\
Cambridge Dictionary & 33.3 & 33.3 & 11.1 & 44.4 & 44.4 & 44.4 & 22.2 & 55.6 & 66.7 & 22.2 & 22.2 & 66.7 & 44.4 & 77.8 & 22.2 & 88.9 & 88.9 & 88.9 & 66.7 & 88.9 \\
Coursera & 50.0 & 100.0 & 0.0 & 100.0 & 50.0 & 50.0 & 0.0 & 50.0 & 50.0 & 0.0 & 0.0 & 50.0 & 50.0 & 100.0 & 0.0 & 50.0 & 100.0 & 100.0 & 100.0 & 100.0 \\
ESPN & 0.0 & 20.0 & 30.0 & 30.0 & 20.0 & 10.0 & 30.0 & 40.0 & 30.0 & 30.0 & 30.0 & 50.0 & 30.0 & 40.0 & 30.0 & 50.0 & 70.0 & 70.0 & 50.0 & 70.0 \\
Google Map & 0.0 & 0.0 & 0.0 & 0.0 & 11.1 & 0.0 & 0.0 & 0.0 & 0.0 & 0.0 & 0.0 & 0.0 & 0.0 & 44.4 & 0.0 & 33.3 & 55.6 & 66.7 & 55.6 & 77.8 \\
Google Search & 25.0 & 37.5 & 25.0 & 37.5 & 18.8 & 37.5 & 25.0 & 50.0 & 12.5 & 43.8 & 25.0 & 50.0 & 18.8 & 31.2 & 25.0 & 50.0 & 18.8 & 50.0 & 43.8 & 68.8 \\
Huggingface & 11.8 & 17.6 & 23.5 & 17.6 & 5.9 & 11.8 & 23.5 & 23.5 & 11.8 & 29.4 & 23.5 & 11.8 & 29.4 & 23.5 & 35.3 & 11.8 & 41.2 & 52.9 & 58.8 & 41.2 \\
Wolfram Alpha & 17.6 & 17.6 & 17.6 & 17.6 & 26.5 & 20.6 & 17.6 & 20.6 & 17.6 & 20.6 & 20.6 & 23.5 & 29.4 & 17.6 & 20.6 & 26.5 & 61.8 & 55.9 & 41.2 & 61.8 \\
\midrule
\textbf{Overall} & 17.8 & 20.2 & 18.6 & \textbf{25.6} & 20.9 & 20.9 & 19.4 & \textbf{27.1} & 21.7 & 24.0 & 20.9 & \textbf{30.2} & 27.1 & 32.6 & 23.3 & \textbf{38.0} & 56.6 & 59.7 & 46.5 & \textbf{63.6} \\
\bottomrule
\end{tabular}
}
\caption{Full comparison of the success rate (SR) of \method{} with its four variants on WebVoyager. R refers to \method{}$_\text{Random}$, G refers to \method{}$_\text{Google}$, MCTS refers to \method{}$_\text{MCTS}$, and M refers to \method{}.}
\label{tab:ablation_webvoyager}
\end{table*}

\begin{table*}[t]
\centering
\small
\begin{tabular}{ll|cccc|cccc|c}
\toprule
& & \multicolumn{4}{c|}{{Single-source QA}} & \multicolumn{4}{c|}{{Multi-source QA}} & {Overall} \\
{Model} & {Method} & Easy & Medium & Hard & Overall & Easy & Medium & Hard & Overall & \\
\midrule
\headerblue \multicolumn{11}{c}{\textit{Open-Sourced LLMs}} \\
\midrule
\multirow{4}{*}{{Qwen3-4B}}
& \method{} $_{\text{random}}$ & 18.75 & 10.71 & 7.50 & 11.47 & 13.75 & 7.14 & 7.50 & 8.82 & 10.15 \\
& \method{} $_{\text{google}}$ & 16.25 & 25.00 & 10.83 & 17.94 & 12.50 & 12.86 & 14.17 & 13.24 & 15.59 \\
& \method{} $_{\text{MCTS}}$ & 15.00 & 10.71 & 12.50 & 12.35 & 7.50 & 14.29 & 10.83 & 11.47 & 11.91 \\
& \method{} & 21.25 & 26.43 & 17.50 & 22.06 & 11.25 & 12.14 & 12.50 & 12.06 & \textbf{17.06} \\
\midrule
\multirow{4}{*}{{Qwen3-8B}}
& \method{} $_{\text{random}}$ & 22.50 & 15.71 & 10.00 & 15.29 & 20.00 & 12.14 & 8.33 & 12.65 & 13.97 \\
& \method{} $_{\text{google}}$ & 23.75 & 25.71 & 16.67 & 22.06 & 21.25 & 14.29 & 10.83 & 14.71 & 18.38 \\
& \method{} $_{\text{MCTS}}$ & 17.50 & 12.86 & 12.50 & 13.82 & 10.00 & 16.43 & 11.67 & 13.24 & 13.53 \\
& \method{} & 26.25 & 31.43 & 25.83 & 28.24 & 16.25 & 15.71 & 15.00 & 15.59 & \textbf{21.91} \\
\midrule
\multirow{4}{*}{{Qwen3-14B}}
& \method{} $_{\text{random}}$ & 27.50 & 25.71 & 10.00 & 20.59 & 21.25 & 11.43 & 10.83 & 13.53 & 17.06 \\
& \method{} $_{\text{google}}$ & 32.50 & 35.71 & 17.50 & 28.53 & 18.75 & 20.71 & 13.33 & 17.65 & 23.09 \\
& \method{} $_{\text{MCTS}}$ & 18.75 & 13.57 & 17.50 & 16.18 & 11.25 & 20.00 & 12.50 & 15.29 & 15.74 \\
& \method{} & 35.00 & 37.14 & 25.83 & 32.65 & 20.00 & 19.29 & 19.17 & 19.41 & \textbf{26.03} \\
\midrule
\multirow{4}{*}{{Qwen3-32B}}
& \method{} $_{\text{random}}$ & 28.75 & 22.14 & 19.17 & 22.65 & 21.25 & 16.43 & 17.50 & 17.94 & 20.29 \\
& \method{} $_{\text{google}}$ & 35.00 & 41.43 & 20.83 & 32.65 & 22.50 & 21.43 & 14.17 & 19.12 & 25.88 \\
& \method{} $_{\text{MCTS}}$ & 18.75 & 14.29 & 18.33 & 16.76 & 11.25 & 22.14 & 12.50 & 16.18 & 16.47 \\
& \method{} & 41.25 & 40.71 & 25.00 & 35.29 & 25.00 & 22.14 & 18.33 & 21.47 & \textbf{28.38} \\
\midrule
\headergreen \multicolumn{11}{c}{\textit{Closed-Sourced LLMs}} \\
\midrule
\multirow{4}{*}{{GPT-5-mini}}
& \method{} $_{\text{random}}$ & 57.50 & 60.71 & 46.67 & 55.00 & 45.00 & 38.57 & 38.33 & 40.00 & 47.50 \\
& \method{} $_{\text{google}}$ & 53.75 & 60.00 & 50.83 & 55.29 & 46.25 & 40.00 & 45.83 & 43.53 & 49.41 \\
& \method{} $_{\text{MCTS}}$ & 50.00 & 41.43 & 44.17 & 44.41 & 40.00 & 41.43 & 38.33 & 40.00 & 42.21 \\
& \method{} & 63.75 & 64.29 & 54.17 & 60.59 & 43.75 & 50.71 & 37.50 & 44.41 & \textbf{52.50} \\
\bottomrule
\end{tabular}
\caption{Full comparison of the success rate (SR) of \method{} with four variants on WebWalkerQA.}
\label{tab:ablation_webwalker_full}
\end{table*}

We report the full experimental results on WebVoyager, categorized by website type in Table~\ref{tab:webvoyager_results_full}. We also report the detailed results of the ablation study on WebVoyager and WebWalkerQA in Table~\ref{tab:ablation_webvoyager} and Table~\ref{tab:ablation_webwalker_full}, respectively.

\end{document}